\documentclass{article} 
\usepackage{iclr2016_conference,times}
\usepackage{url}
\usepackage{algorithm}
\usepackage{algpseudocode}
\usepackage{graphicx}
\usepackage{amsfonts}
\usepackage{amsmath}
\usepackage{natbib}
\usepackage{verbatim}
\usepackage{booktabs}
\usepackage[rightcaption]{sidecap}
\usepackage[hidelinks]{hyperref}

\title{Efficient Parallel Methods for Deep Reinforcement Learning}

\author{
Alfredo V. Clemente \\
Department of Computer and Information Science\\
Norwegian University of Science and Technology\\
Trondheim, Norway \\
\texttt{alfredvc@stud.ntnu.no} \\
\And
Humberto N. Castej\'{o}n\\
Telenor Research\\
Trondheim, Norway \\
\texttt{humberto.castejon@telenor.com} \\
\AND
Arjun Chandra \\
Telenor Research \\
Trondheim, Norway \\
\texttt{arjun.chandra@telenor.com} \\
}

\newcommand{\vect}[1]{\boldsymbol{#1}}

\algblock{ParFor}{EndParFor}
\algnewcommand\algorithmicparfor{\textbf{parallel for}}
\algnewcommand\algorithmicpardo{\textbf{do}}
\algnewcommand\algorithmicendparfor{\textbf{end\ parallel for}}
\algrenewtext{ParFor}[1]{\algorithmicparfor\ #1\ \algorithmicpardo}
\algrenewtext{EndParFor}{\algorithmicendparfor}

\iclrfinalcopy 

\begin{document}

\maketitle

\begin{abstract}
We propose a novel framework for efficient parallelization of deep reinforcement learning algorithms, enabling these algorithms to learn from multiple actors on a single machine. The framework is algorithm agnostic and can be applied to on-policy, off-policy, value based and policy gradient based algorithms. Given its inherent parallelism, the framework can be efficiently implemented on a GPU, allowing the usage of powerful models
while significantly reducing training time. We demonstrate the effectiveness of our framework by implementing an advantage actor-critic algorithm on a GPU, using on-policy experiences and employing synchronous updates. Our algorithm achieves state-of-the-art performance on the Atari domain after only a few hours of training. Our framework thus opens the door for much faster experimentation on demanding problem domains. Our implementation is open-source and is made public at \url{https://github.com/alfredvc/paac}.
\end{abstract}

\section{Introduction and Related Work}

Incorporating deep learning models within reinforcement learning (RL) presents some challenges. Standard stochastic gradient descent algorithms for training deep learning models assume all training examples to be 
independent and identically distributed (i.i.d.). This constraint is often violated by RL agents, given the high 
correlation between encountered states. Additionally, when learning on-policy, the policy affects the distribution of encountered states, which in turn affects the policy, creating a feedback loop that may lead to divergence~\citep{atari_rl_2013}. Two main attempts have been made to solve these issues. One is to store experiences in a large replay memory and employ off-policy RL methods~\citep{atari_rl_2013}. Sampling the replay memory can break 
the state correlations, thus reducing the effect of the feedback loop. Another is to execute multiple asynchronous agents in parallel, each interacting with an instance of the environment independently of each other~\citep{a3c}. Both of these approaches suffer from different drawbacks; experience replay can only be employed with off-policy methods, while asynchronous agents can perform inconsistent parameter updates due to stale gradients\footnote{Gradients may be computed w.r.t. stale parameters while updates applied to a new parameter set.} and simultaneous parameter updates from different threads.

Parallel and distributed compute architectures have motivated innovative modifications 
to existing RL algorithms to efficiently make use of parallel execution. 
In the General Reinforcement Learning Architecture (Gorila)~\citep{gorilla}, the DQN~\citep{mnih2015nature} algorithm is distributed across multiple machines. 
Multiple \textit{learners} learn off-policy using experiences collected into a common replay memory by multiple \textit{actors}. Gorila is shown to outperform standard DQN in a variety of Atari games, while only training for 4 days. The distribution of the learning process is further explored in~\citep{a3c}, where multiple actor-learners are executed asynchronously on a single machine. Each actor-learner holds its own copy of the policy/value function and interacts with its own instance of the environment. This allows for both off-policy, as well as on-policy learning.  The actor-learners compute gradients in parallel and update shared parameters asynchronously in a \textsc{Hogwild!}~\citep{hogwild} fashion. The authors suggest that multiple agents collecting independent 
experiences from their own environment instances reduces correlation between samples, 
thereby improving learning. The asynchronous advantage actor-critic (A3C) algorithm~\citep{a3c} was able to 
surpass the state of the art on the Atari domain at the time of publication, 
while training for 4 days on a single machine with 16 CPU cores. GA3C~\citep{nvidia} is a GPU implementation of A3C. It batches action selection and learning using queues. Actors sample from a shared policy by submitting a task to a \textit{predictor}, which executes the policy and returns an action once it has accumulated enough tasks. Similarly, learning is performed by submitting experiences to a \textit{trainer}, which computes gradients and applies updates once enough experiences have been gathered. If the training queue is not empty when the model is updated, the learning will no longer be on-policy, since the remaining experiences were generated by an old policy. This leads to instabilities during training, which the authors address with a slight modification to the weight updates.

We propose a novel framework for efficient parallelization of deep reinforcement learning algorithms, which keeps the strengths of the aforementioned approaches, while alleviating their weaknesses. Algorithms based on this framework can learn from hundreds of actors in parallel, similar to Gorila, while running on a single machine like A3C and GA3C.  Having multiple actors help decorrelate encountered states and attenuate feedback loops, while allowing us to leverage the parallel architecture of modern CPUs and GPUs. Unlike A3C and Gorila, there is only one copy of the parameters, hence parameter updates are performed synchronously, thus avoiding the possible drawbacks related to asynchronous updates. Our framework has many similarities to GA3C. However, the absence of queues allows for a much more simpler and computationally efficient solution, while allowing for true on-policy learning and faster convergence to optimal policies. We demonstrate our framework with a Parallel Advantage Actor-Critic algorithm, that achieves state of the art performance in the Atari 2600 domain after only a few hours of training. This opens the door for much faster experimentation.
 
\section{Background}
Reinforcement learning algorithms attempt to learn a policy $\pi$ that maps states to actions, in order to maximize the expected sum of cumulative rewards $R_t = \mathbb{E}_{\pi}\big[\sum_{k=0}^{\infty}\gamma^kr_{t+k}\big]$ for some discount factor $0 < \gamma < 1$, where $r_t$ is the reward observed at timestep $t$. Current reinforcement learning algorithms represent the learned policy $\pi$ as a neural network, either implicitly with a value function or explicitly as a policy function.

\subsection{Batching with Stochastic Gradient Descent}
Current reinforcement learning algorithms make heavy use of deep neural networks, both to extract high level features from the observations it receives, and as function approximators to represent its policy or value functions. \\
Consider the set of input-target pairs $S=\{(x_0,y_0),(x_1, y_1),...(x_n,y_n)\}$ generated by some function $f^*(x)$. The goal of supervised learning with neural networks is to learn a parametrized function $f(x;\theta)$ that best approximates function $f*$. The performance of $f$ is evaluated with the empirical loss
\begin{equation}
L(\theta) = \frac{1}{|S|}\sum\limits_{s\in S}l(f(x_s;\theta), y_s)
\end{equation}
where $l(f(x_s;\theta), y_s)$ is referred to as a loss function, and gives a quantitative measure of how good $f$ is at modelling $f^*$. The model parameters $\theta$ are learned with stochastic gradient descent (SGD) by iteratively applying the update 
$$
\theta_{i+1} \gets \theta_i - \alpha\nabla_{\theta_i}L(\theta_i)
$$

for a learning rate $\alpha$. In SGD, $L(\theta)$ is usually approximated with 
$$\bar{L}(\theta)=\frac{1}{|S'|}\sum_{s'\in S'}l(f(x_{s'};\theta), y_{s'}),$$
where
$S'\subset S$ is a mini-batch sampled from $S$. The choice of $\alpha$ and $n_{S'}=|S'|$ presents a trade-off between computational efficiency and sample efficiency. Increasing $n_{S'}$ by a factor of $k$ increases the time needed to calculate $\nabla_{\theta}\bar{L}$ by a factor of $k'$, for $k' \leq k$, and reduces its variance proportionally to $\frac{1}{k}$~\citep{optimization_for_large_scale}. In order to mitigate the increased time per parameter update we can increase the learning rate to $\alpha'$ for some $\alpha' \geq \alpha$. However there are some limits on the size of the learning rate, so that in general $\alpha' \leq k\alpha$~\citep{optimization_for_large_scale}. The hyper-parameters $\alpha'$ and $k$ are chosen to simultaneously maximize $\frac{k}{k'}$ and minimize $L$.

\subsection{Model-free Value Based Methods}
Model-free value based methods attempt to model the Q-funciton $q(s_t, a_t)=\mathbb{E}_{\pi}\big[ \sum_{k=0}^{\infty}\gamma^kr_{t+k+1}\big| s=s_t, a=a_t \big]$ that gives the expected return achieved by being in state $s_t$ taking action $a_t$ and then following the policy $\pi$. A policy can be extracted from the Q-function with $\pi(s_t)=\arg\max_{a'}q(s_t,a')$. DQN~\citep{atari_rl_2013} learns a function $Q(s,a;\theta)\approx q(s,a)$ represented as a convolutional neural network with parameters $\theta$. Model parameters are updated based on model targets provided by the Bellman equation
\begin{equation}
q(s,a)=\mathbb{E}_{s'}\Big[r + \gamma \max_{a'}q(s',a')|s,a\Big]
\end{equation}
to create the loss function 
\begin{equation}
\bar{L}(\theta_{i}) = (r_t + \gamma \max_{a'}Q(s_{t+1},a';\theta_i) - Q(s_t, a_t;\theta_i))^2
\end{equation}

The parameters $\theta$ are improved by SGD with the gradient

\begin{equation}
\label{eq:dqntarget}
\nabla_{\theta_i}L(\theta_i)\approx\nabla_{\theta_i}\bar{L}(\theta_{i})=-\frac{1}{2}\big( r + \gamma \max_{a'}Q(s',a';\theta_{i})-Q(s,a;\theta_i) \big) \nabla_{\theta_i}Q(s,a;\theta_i)
\end{equation}

\subsection{Policy Gradient Methods}
\label{subsed:policy_gradient_methods}
Policy gradient methods~\citep{policy_gradient} directly learn a parametrized policy $\pi(a|s;\theta)$. This is possible due to the policy gradient theorem~\citep{sutton1999policy}
\begin{equation}
\label{eq:policy_gradient_loss}
\nabla_{\theta}L(\theta)=\mathbb{E}_{s,a}\big[q(s,a) \nabla_{\theta}\log\pi(a|s;\theta) \big],
\end{equation}
which provides an unbiased estimate of the gradient of the return with respect to the policy parameters.
\citet{sutton1999policy} propose an improvement upon the basic policy gradient update by replacing the Q function with the advantage function $A_{\pi}(s,a)=(q(s,a) - v(s))$ where $v(s)$ is the value function given by $\mathbb{E}_{\pi}\big[ \sum_{k=0}^{\infty}\gamma^kr_{t+k+1}\big| s=s_t\big]$. When $\pi$ is continuously differentiable, $\theta$ can be optimized via gradient ascent following

\begin{equation}
\label{eq_policy_loss_exp}
\nabla_{\theta}L(\theta)=\mathbb{E}_{s,a}\big[ \big( q(s,a) - v(s) \big) \nabla_{\theta}\log\pi(a|s;\theta) \big]
\end{equation}

\citet{a3c} learn an estimate $V(s;\theta_v)\approx v(s)$ of the value function, with both $V(s;\theta_v)$ and $\pi(a|s;\theta)$ being represented as convolutional neural networks. Additionally, they estimate the Q-function with the n-step return estimate given by 

\begin{equation}
Q^{(n)}(s_t,a_t;\theta,\theta_v) = r_{t+1}+...+\gamma^{n-1}r_{t+n-1} + \gamma^nV(s_{t+n};\theta_v)
\end{equation}

with $0 < n \leq t_{max}$ for some fixed $t_{max}$. The final gradient for the policy network is given by

\begin{equation}
\label{eq_policy_loss}
\nabla_{\theta}L(\theta)\approx\big( Q^{(n)}(s_t,a_t;\theta,\theta_v) - V(s_t;\theta_v) \big) \nabla_{\theta}\log\pi(a_t|s_t;\theta)
\end{equation}

The gradient for the value network $V$ is given by
\begin{equation}
\label{eq:value_loss}
\nabla_{\theta_v}L(\theta_v) \approx \nabla_{\theta_v}\Big[\big( Q^{(n)}(s_t,a_t;\theta,\theta_v) - V(s_t;\theta_v) \big)^2\Big]
\end{equation}

Samples are generated by having an actor interact with an environment by following the policy $\pi(s_t)$ and observing the next state $s_{t+1}$ and reward $r_{t+1}$.

\section{Parallel Framework for Deep Reinforcement Learning}
\label{parallel_framework}
\begin{figure}[htb]
  \centering
  \includegraphics[width=0.8\linewidth]{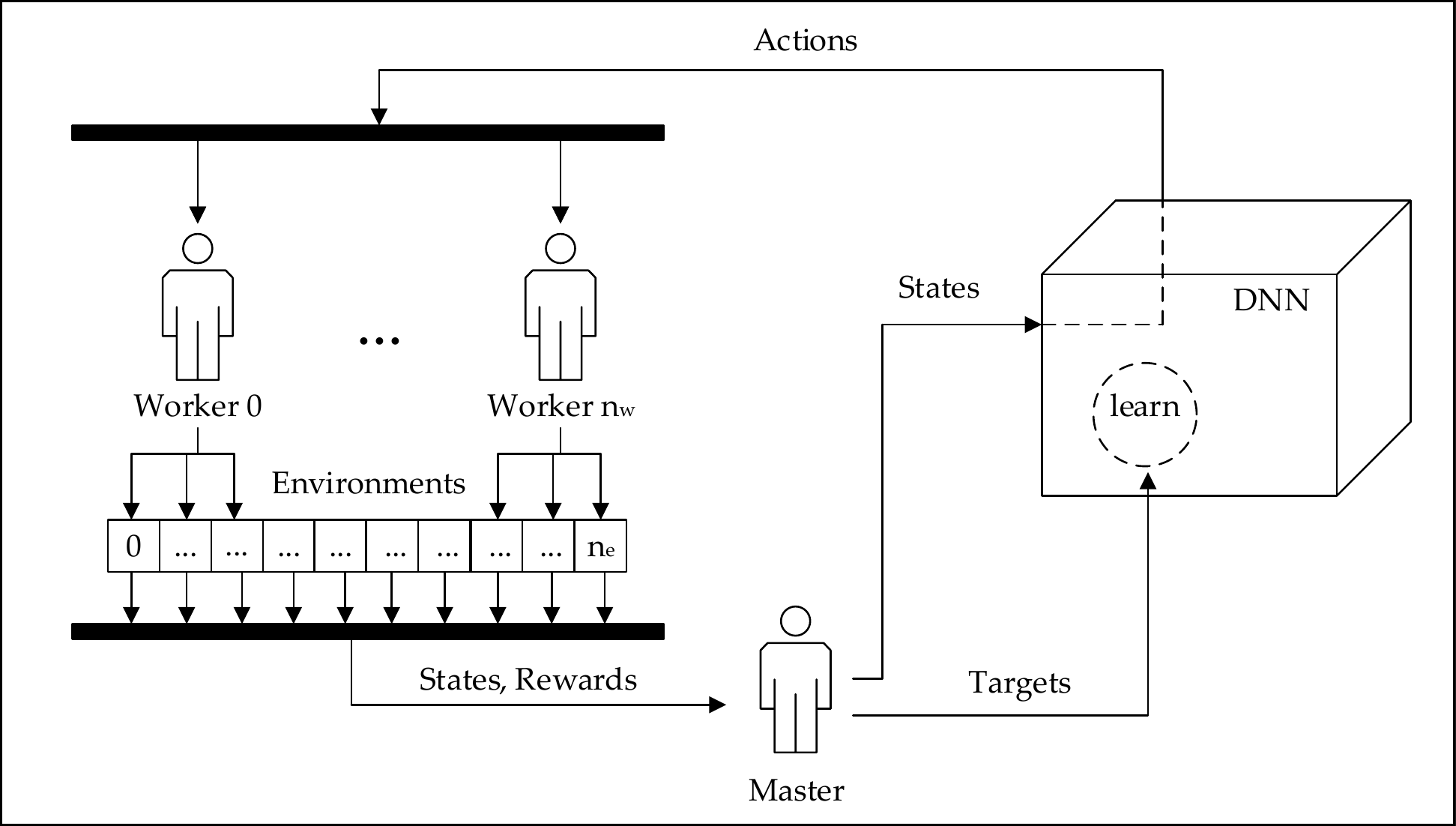}
  \caption{Architecture of the Parallel Framework for Deep RL}
  \label{fig:para_frameworl}
\end{figure}

We propose a general framework for deep reinforcement learning, where multiple actors can be trained synchronously on a single machine. A set of $n_e$ environment instances are maintained, where actions for all environment instances are generated from the policy. The architecture can be seen in Figure~\ref{fig:para_frameworl}.
The policy function can be represented implicitly, as in value based methods, or explicitly as in policy gradient methods. As suggested in \citet{a3c}, by having multiple environments instances in parallel it is likely that they will be exploring different locations of the state space at any given time, which reduces the correlation of encountered states and helps stabilize training. This approach can be motivated as an \textit{on-line} experience memory, where experiences are sampled from the distribution currently being observed from the environment, instead of sampling uniformly from previous experience.

At each timestep the master generates actions for all environment instances by sampling the current policy. Note, that the policy may be sampled differently for each environment. A set of $n_w$ workers then apply all the actions to the their respective environments in parallel, and store the observed experiences. The master then updates the policy with the observed experiences. This allows the evaluation and training of the policy to be batched, which can be efficiently parallelized, leading to significant speed improvements on modern compute architectures.

\section{Parallel Advantage Actor Critic}
\label{paac}
We used the proposed framework to implement a version of the n-step advantage actor-critic algorithm
proposed by \citet{a3c}. This algorithm maintains a policy $\pi(a_{t}|s_{t};\theta)$
and an estimate $V(s_{t};\theta_{v})$ of the value function, both
approximated by deep neural networks. The parameters $\theta$ of
the policy network (the actor) are optimized via gradient ascent following $$\nabla_{\theta}\log\pi(a_{t}|s_{t};\theta)A(s_{t},a_{t};\theta,\theta_{v})+\beta\nabla_{\theta}H(\pi(s_{e,t};\theta))$$
~\citep{sutton1999policy}, where $A(s_{t},a_{t};\theta,\theta_{v})=Q^{(n)}(s_{t},a_{t};\theta,\theta_{v})-V(s_{t};\theta_{v})$
is an estimate of the advantage function, $Q^{(n)}(s_{t},a_{t};\theta,\theta_{v})=\sum_{k=0}^{n-1}\gamma^{k}r_{t+k}+\gamma^{n}V(s_{t+n};\theta_{v})$, with $0<n\leq t_{max}$, is the \emph{n}-step return estimation and $H(\pi(s_{e,t};\theta))$ is the entropy of the policy $\pi$, which as suggested by \citet{a3c} is added to improve exploration by discouraging premature convergence to suboptimal deterministic policies. The
parameters $\theta_{v}$ of value network (the critic) are in turn
updated via gradient descent in the direction of
$$\nabla_{\theta_{v}}\Big[\big(Q^{(n)}(s_{t},a_{t};\theta,\theta_{v})-V(s_{t};\theta_{v})\big)^{2}\Big]$$

In the context of our framework, the above gradients are calculated
using mini batches of experiences. For each of the $n_{e}$ environments,
$t_{max}$ experiences are generated, resulting in batches of size
$n_{e}\cdotp t_{max}$. The gradients $\nabla_{\theta}^{\pi}$ for
the policy network and the gradients $\nabla_{\theta}^{V}$ for the
value function thus take the following form:

\begin{equation}
\nabla_{\theta}^{\pi}\approx\frac{1}{n_e\cdot t_{\max}}\sum_{e=1}^{n_e}\sum_{t=1}^{t_{\max}} \big( Q^{(t_{\max}-t+1)}(s_{e,t},a_{e,t};\theta,\theta_v) - V(s_{e,t};\theta_v) \big) \nabla_{\theta}\log\pi(a_{e,t}|s_{e,t};\theta)+\beta\nabla_{\theta}H(\pi(s_{e,t};\theta))
\end{equation}

\begin{equation}
\nabla_{\theta_{v}}^{V}\approx\nabla_{\theta_{v}}\frac{1}{n_{e}\cdot t_{max}}\sum_{e=1}^{n_{e}}\sum_{t=1}^{t_{\max}}\big(Q^{(t_{\max}-t+1)}(s_{e,t},a_{e,t};\theta,\theta_{v})-V(s_{e,t};\theta_{v})\big)^{2}
\end{equation}

Pseudocode for our parallel advantage actor-critic algorithm (PAAC)
is given in Algorithm~\ref{alg:paac}. As shown in the next section,
PAAC achieves state of the art performance on the Atari 2600 domain
in half of the time required by GA3C and in only one eigth of the time
required by A3C. 
Note that, although we implement an actor-critic algorithm, this framework can be used to implement any other reinforcement learning algorithm.

\section{Experiments}
\label{headings}
We evaluated the performance of PAAC in 12 games from Atari 2600 using the Atari Learning Environment~\citep{ale}. The agent was developed in Python using TensorFlow~\citep{abadi2015tensorflow} and all performance experiments were run on a computer with a 4 core Intel i7-4790K processor and an Nvidia GTX 980 Ti GPU.
\subsection{Experimental setup}
\label{exp_setup}
To compare results with other algorithms for the Atari domain we follow the same pre-processing and training procedures as \citet{a3c}. Each action is repeated 4 times, and the per-pixel maximum value from the two latest frames is kept. The frame is then scaled down from $210\times 160$ pixels and 3 color channels to $84\times 84$ pixels and a single color channel for pixel intensity. Whenever an environment is restarted, the state is reset to the starting state and between 1 and 30 no-op actions are performed before giving control to the agent. The environment is restarted whenever the final state of the environment is reached.

\begin{algorithm}[H]
\small
\caption{Parallel advantage actor-critic}
\label{alg:paac}
\begin{algorithmic}[1]
\State Initialize timestep counter $N = 0$ and network weights $\theta$, $\theta_v$
\State Instantiate set $\vect{e}$ of $n_e$ environments 
\Repeat
\For{$t=1$ to $t_{max}$} 
\State Sample $\vect{a}_t$ from $\pi(\vect{a}_t|\vect{s}_t;\theta)$
\State Calculate $\vect{v}_t$ from $V(\vect{s}_t;\theta_v)$
\ParFor{$i=1$ to $n_e$}\;\label{alg:par_for_start}
\State Perform action $a_{t,i}$ in environment $e_i$
\State Observe new state $s_{t+1,i}$ and reward $r_{t+1,i}$
\EndParFor \;\label{alg:par_for_end}
\EndFor 
\State $\vect{R}_{t_{\max}+1} =
    \left\{
    \begin{array}{l l}
      0  \quad & \text{for terminal } \vect{s}_t\\
      V(s_{t_{\max}+1};\theta) & \text{for non-terminal } \vect{s}_t
    \end{array} \right.$
\For {$t=t_{\max}$ down to $1$}
\State $\vect{R}_t = \vect{r}_t + \gamma \vect{R}_{t+1}$
\EndFor
\State $d\theta=\frac{1}{n_e\cdot t_{max}}\sum_{i=1}^{n_e}\sum_{t=1}^{t_{max}}(R_{t,i} - v_{t,i})\nabla_{\theta}\log\pi(a_{t,i}|s_{t,i};\theta) +\beta\nabla_{\theta}H(\pi(s_{e,t};\theta))$\;\label{alg:grad_start}
\State $d\theta_v=\frac{1}{n_e\cdot t_{max}}\sum_{i=1}^{n_e}\sum_{t=1}^{t_{max}}\nabla_{\theta_v} \left(R_{t,i} - V(s_{t,i};\theta_v)\right)^2$\;\label{alg:grad_end}
\State Update $\theta$ using $d\theta$ and $\theta_v$ using $d\theta_v$.
\State $N \gets N + n_e\cdot t_{\max}$
\Until $N \geq N_{max}$
\end{algorithmic}
\end{algorithm}

As in~\citep{a3c}, a single convolutional network with two separate
output layers was used to jointly model the policy and the value functions.
For the policy function, the output is a softmax with one node per
action, while for the value function a single linear output node is
used. Moreover, to compare the efficiency of PAAC for different model
sizes, we implemented two variants of the policy and value convolutional
network. The first variant, referred to as $\mathcal{\text{arch}_{\text{nips}}}$,
is the same architecture used by A3C~FF~\citep{a3c}, which is a
modified version of the architecture used in \citet{atari_rl_2013},
adapted to an actor-critic algorithm. The second variant, $\text{arch}_{\text{nature}}$,
is an adaptation of the architecture presented in \citet{mnih2015nature}.
The networks were trained with RMSProp. The hyperparameters used to
generate the results in Table~\ref{table:results} were $n_{w}=8$,
$n_{e}=32$, $t_{\max}=5$, $N_{\max}=1.15\times10^{8}$, $\gamma=0.99$,
$\alpha=0.0224$, $\epsilon=0.1$, $\beta=0.01$, and a discount factor
of $0.99$ for RMSProp. Additionally gradient clipping \citep{gradient_clipping} with a threshold of 40 was used.

\subsection{Results}
\label{eval}
The performance of PAAC with $\text{arch}_{\text{nips}}$ and $\text{arch}_{\text{nature}}$ was evaluated on twelve different Atari 2600 games, where agents were trained for 115 million skipped frames (460 million actual frames). The results and their comparison to Gorila~\citep{gorilla}, A3C~\citep{a3c} and GA3C~\citep{nvidia} are presented in Table~\ref{table:results}. After a few hours of training on a single computer, PAAC is able to outperform Gorila in 8 games, and A3C FF in 8 games. Of the 9 games used to test GA3C, PAAC matches its performance in 2 of them and surpasses it in the remaining 7.

\begin{figure}[htp]       
    \mbox{\includegraphics[width=0.32\linewidth]{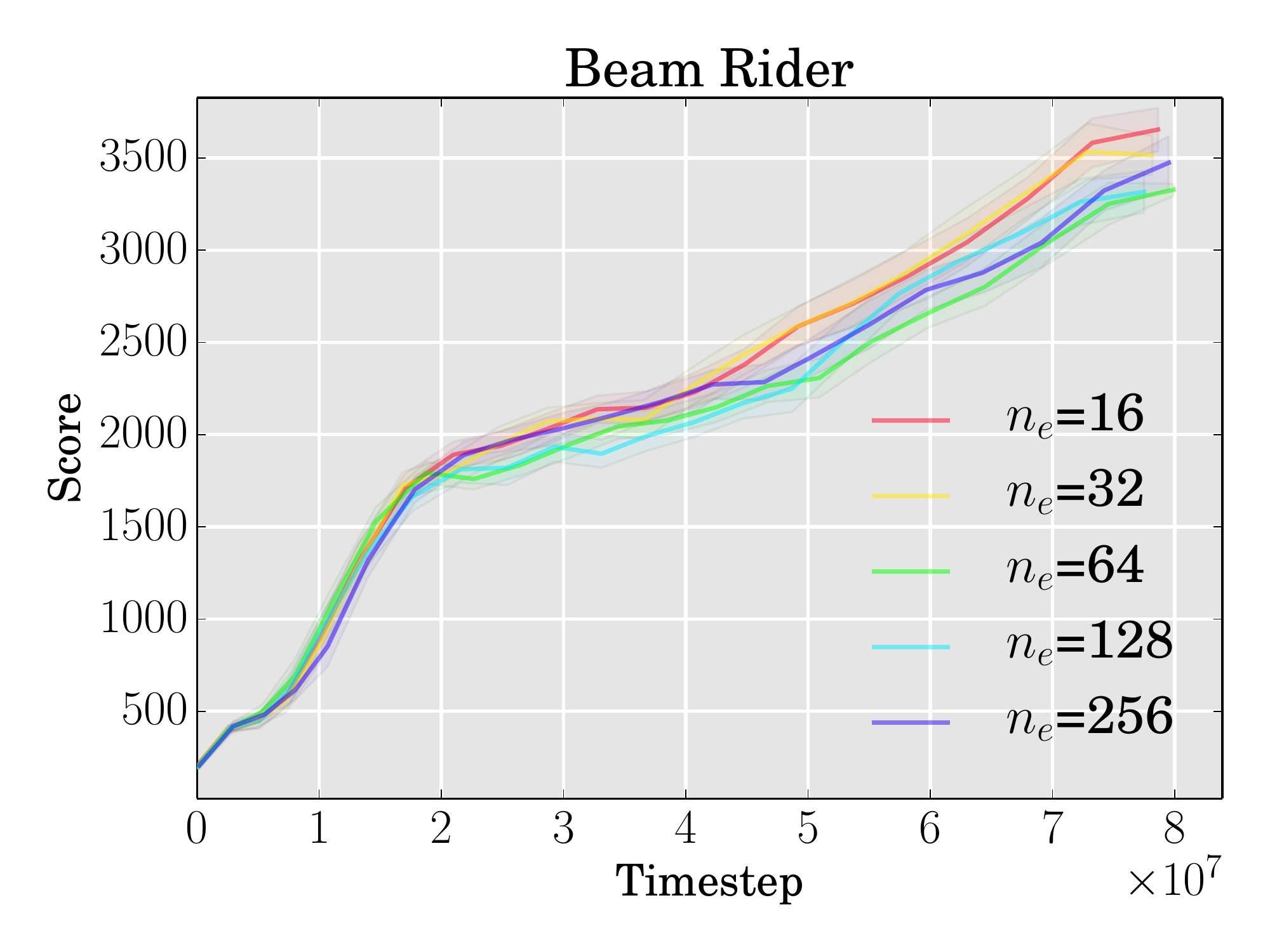}}   
    \hspace{1px}
    \mbox{\includegraphics[width=0.32\linewidth]{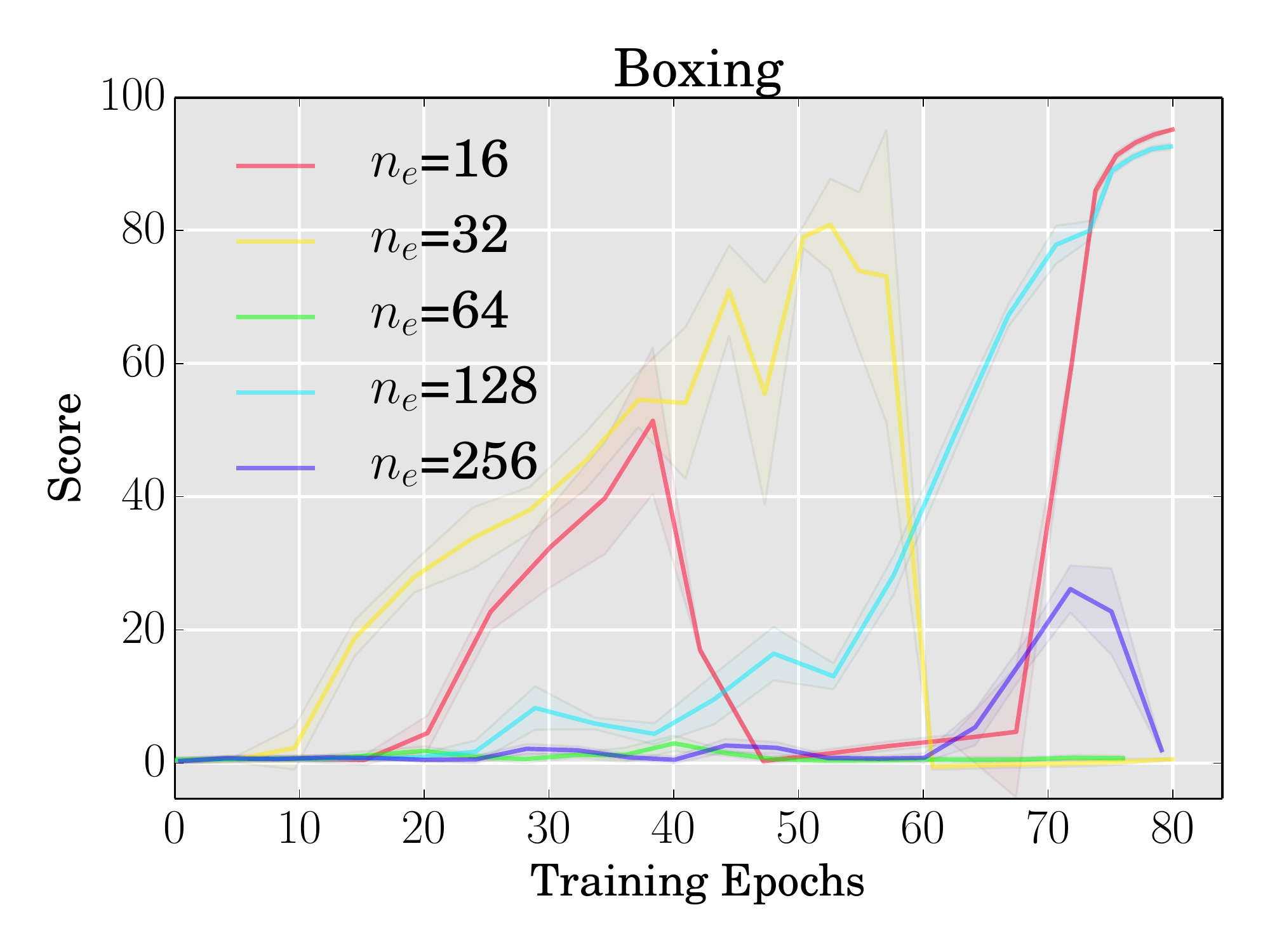}}
    \hspace{1px}
    \mbox{\includegraphics[width=0.32\linewidth]{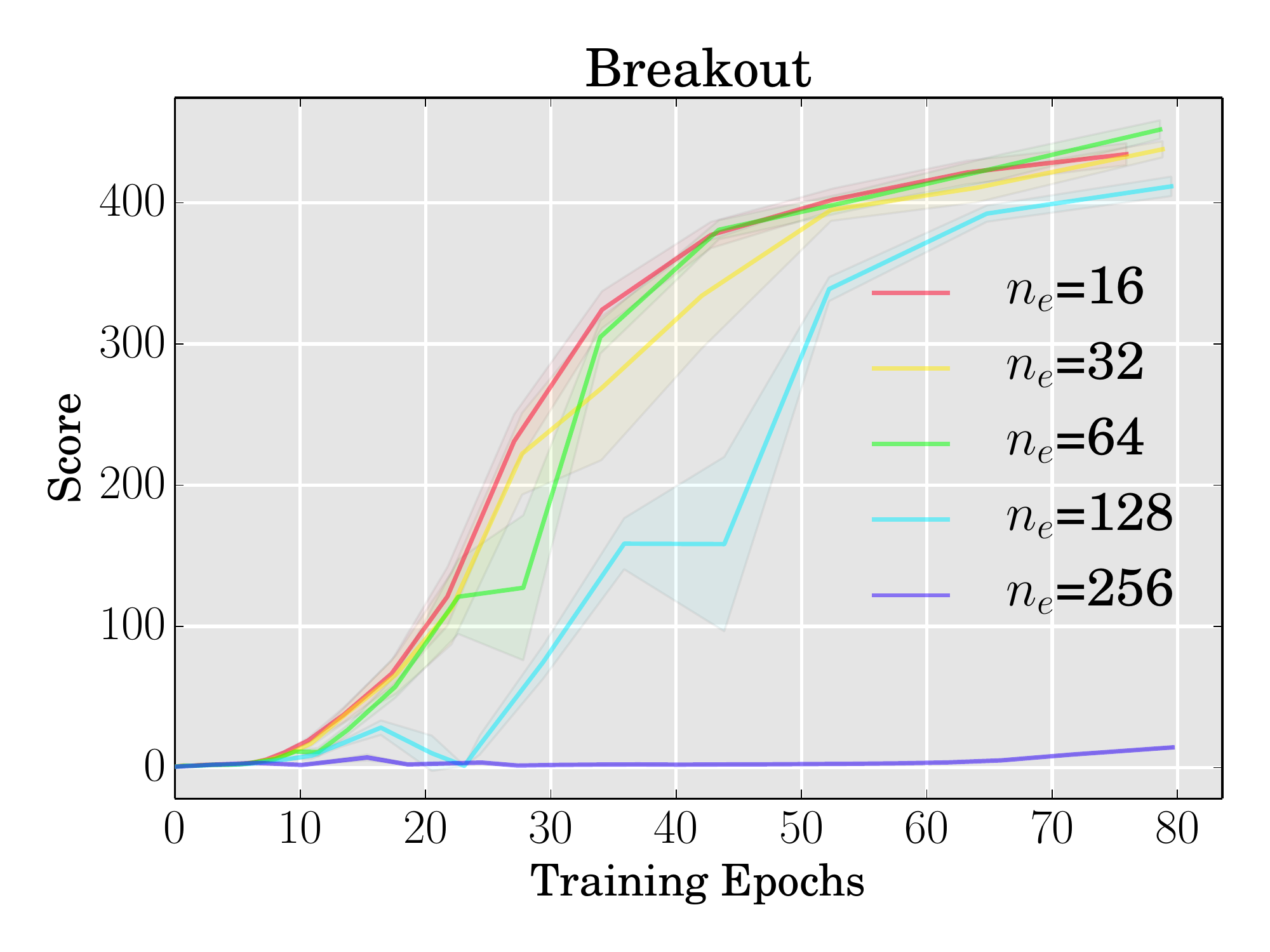}}        
    \mbox{\includegraphics[width=0.32\linewidth]{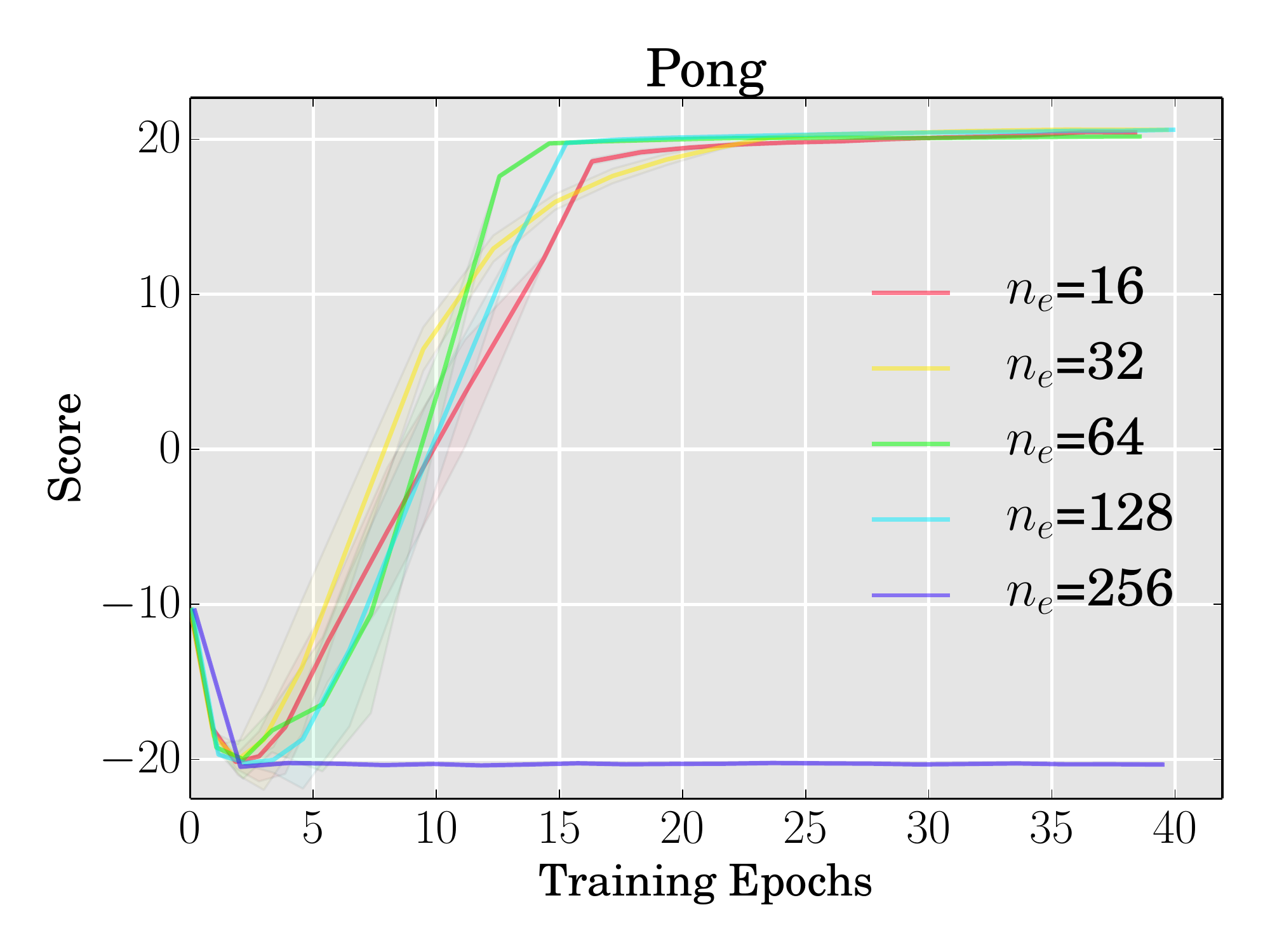}}   
    \hspace{2px}
    \mbox{\includegraphics[width=0.32\linewidth]{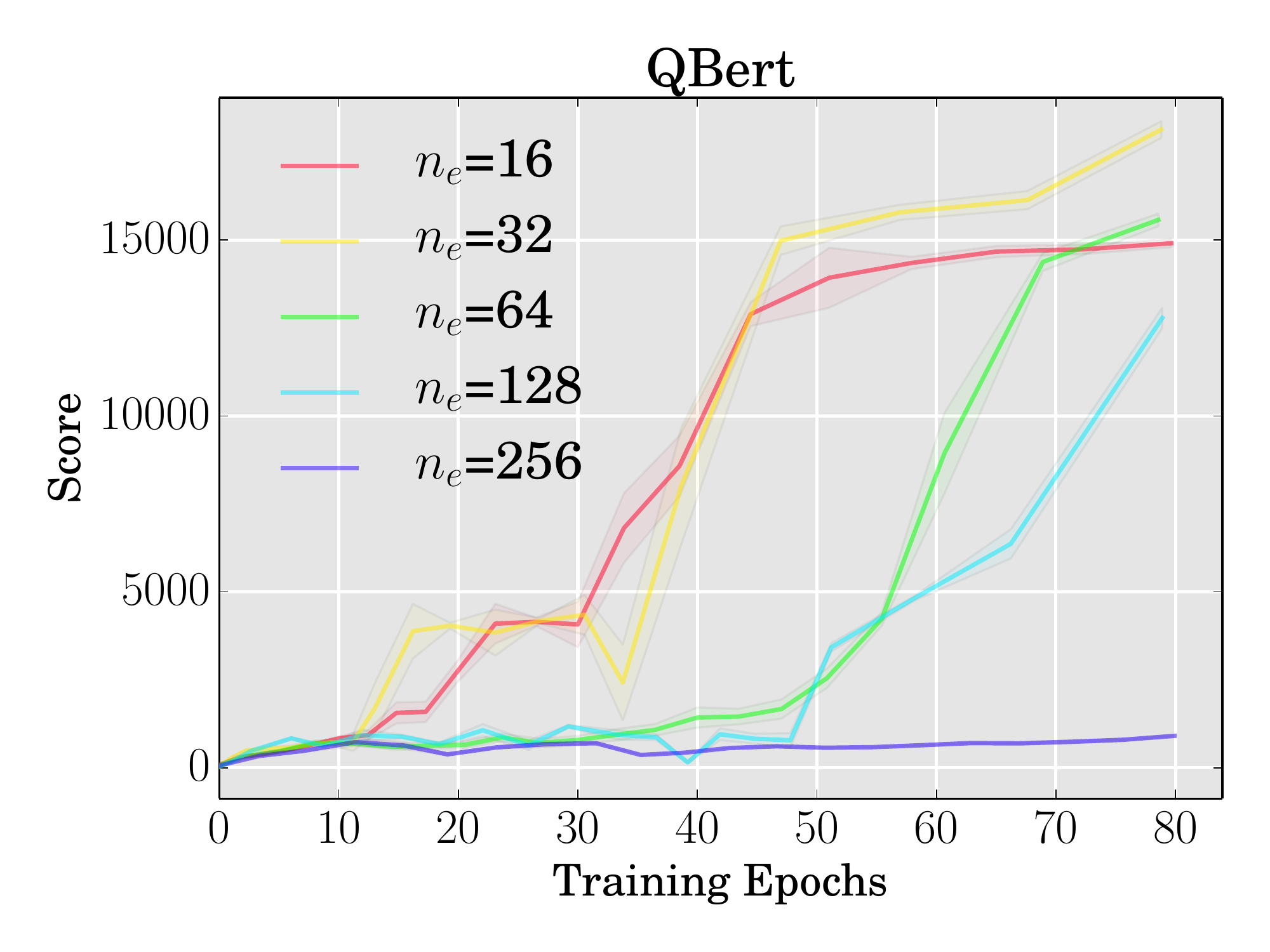}}
    \hspace{1px}
    \mbox{\includegraphics[width=0.32\linewidth]{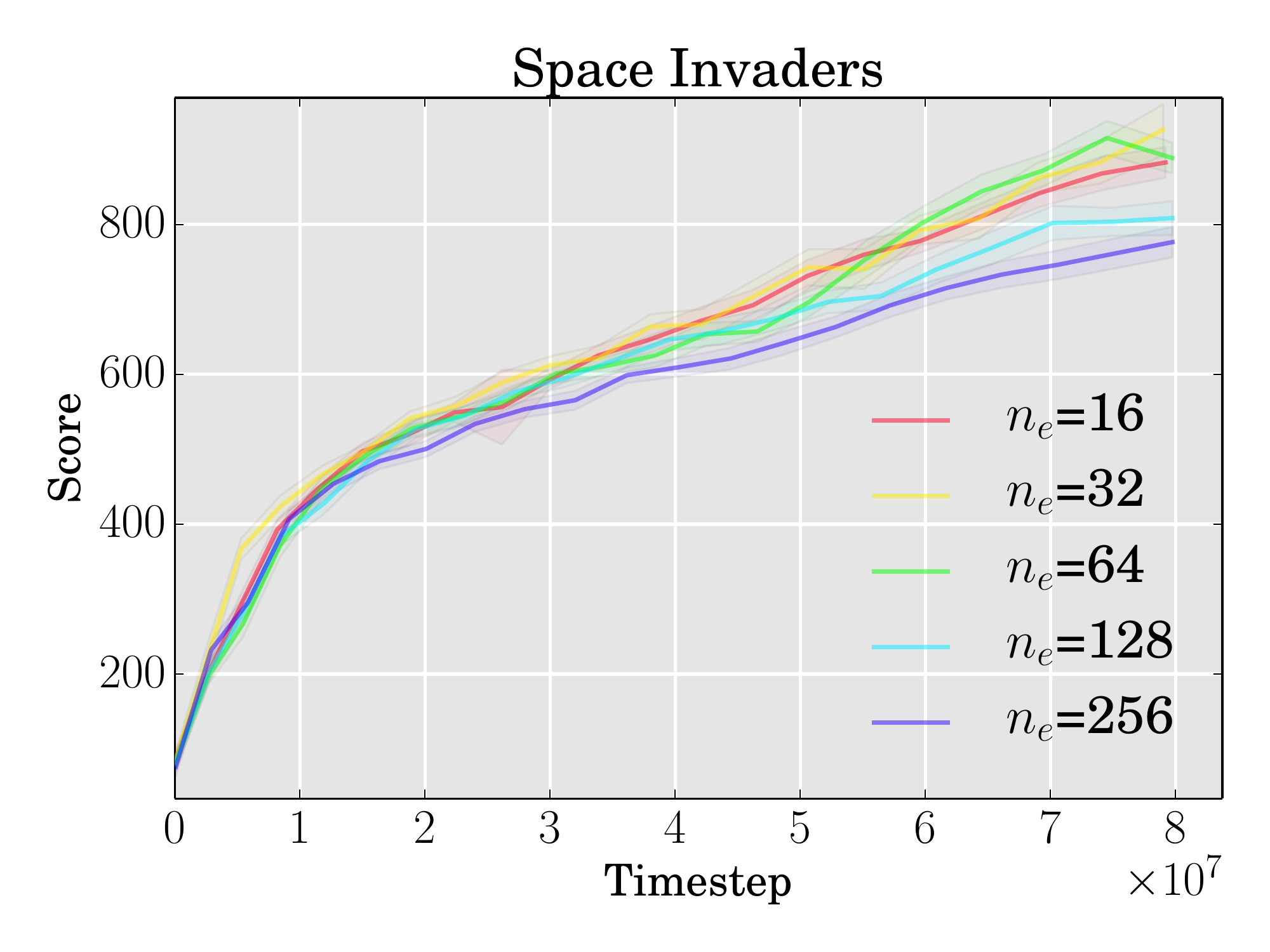}}
    \caption{Score comparison for PAAC on six Atari 2600 games for different $n_e$, where one training epoch is equivalent to 1 million timesteps (4 million skipped frames).}
    \label{fig:score_comparison}
\end{figure}

\begin{figure}[ht]       
    \mbox{\includegraphics[width=0.32\linewidth]{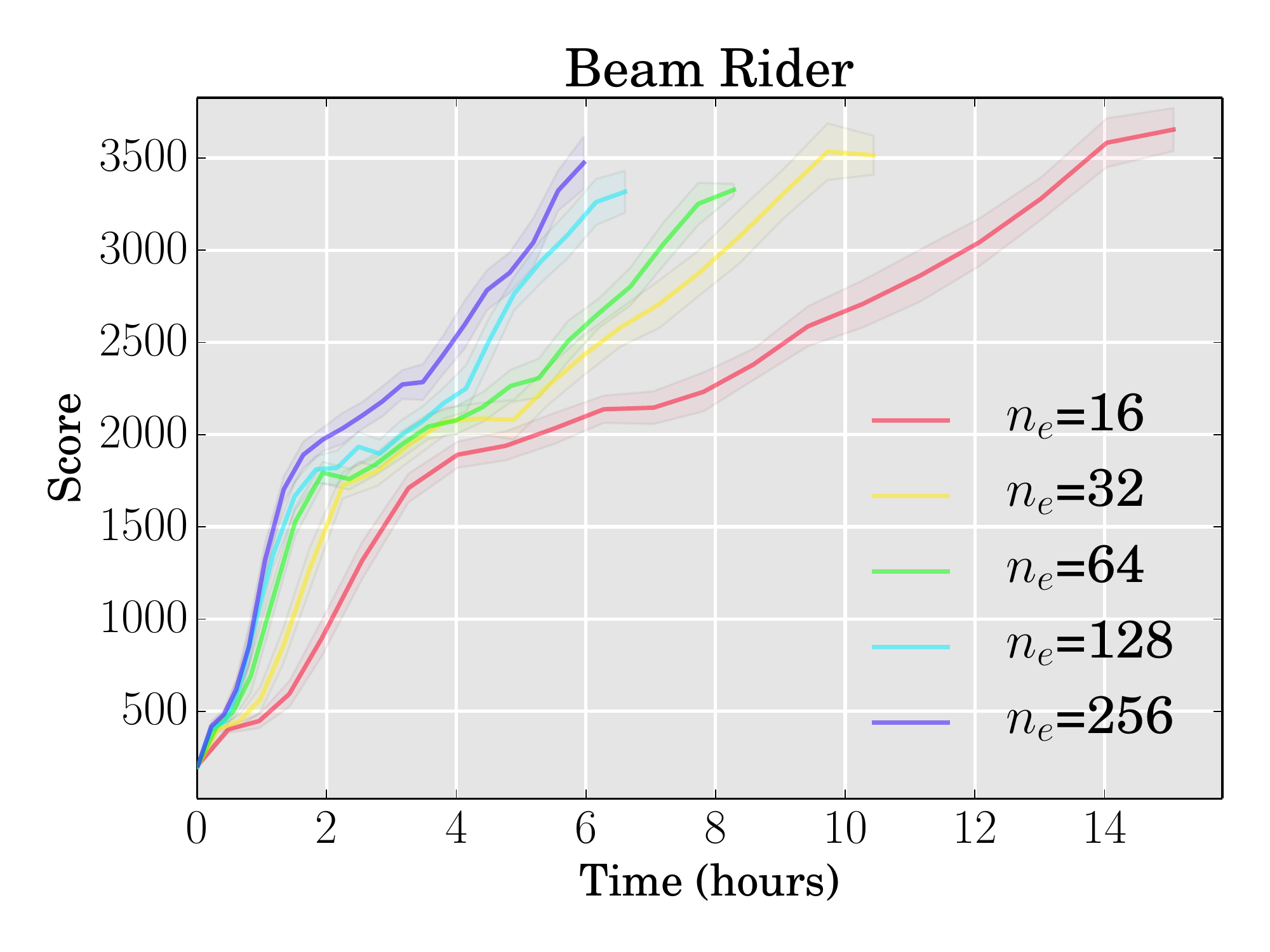}}   
    \hspace{1px}
    \mbox{\includegraphics[width=0.32\linewidth]{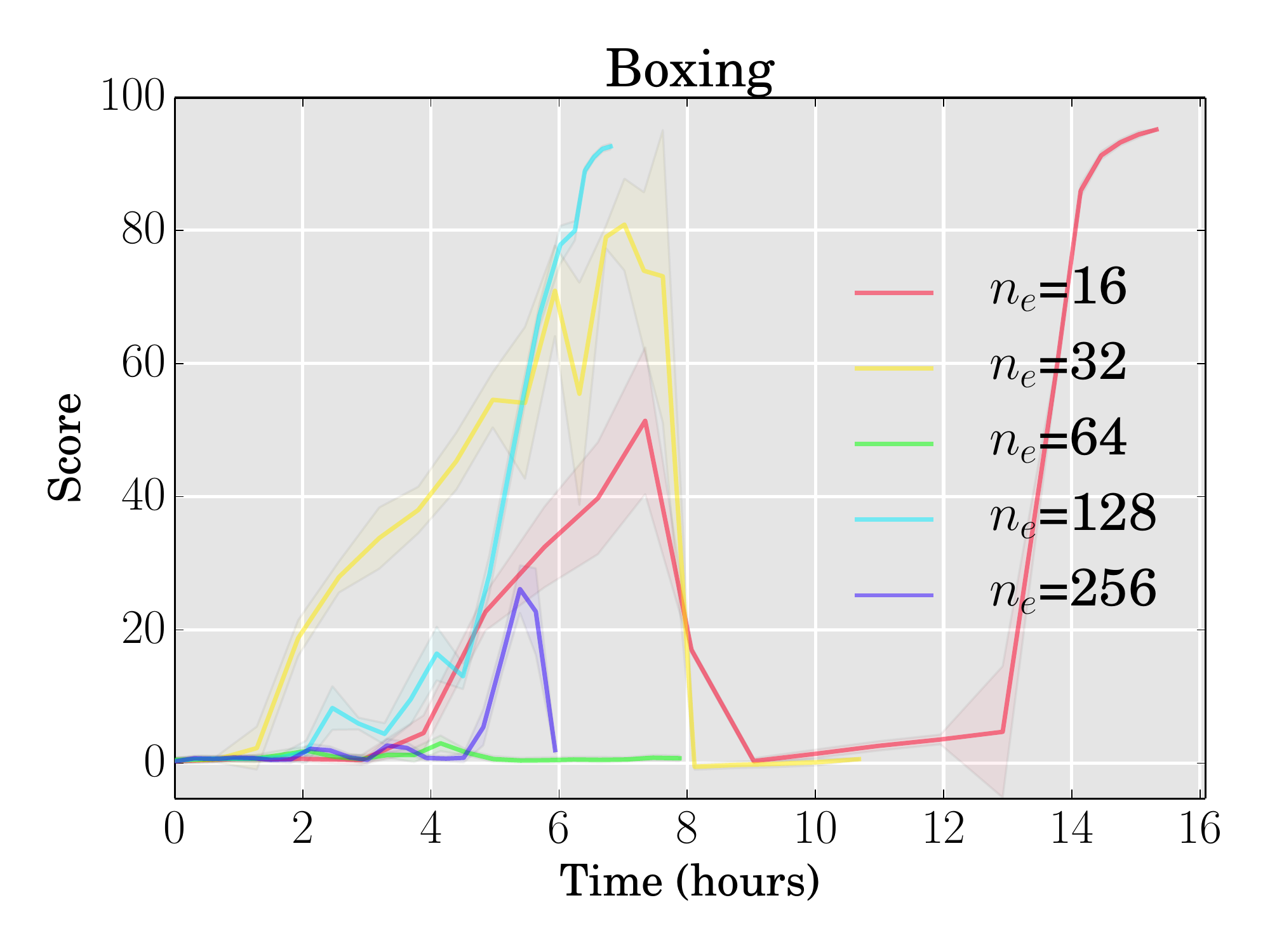}}
    \hspace{1px}
    \mbox{\includegraphics[width=0.32\linewidth]{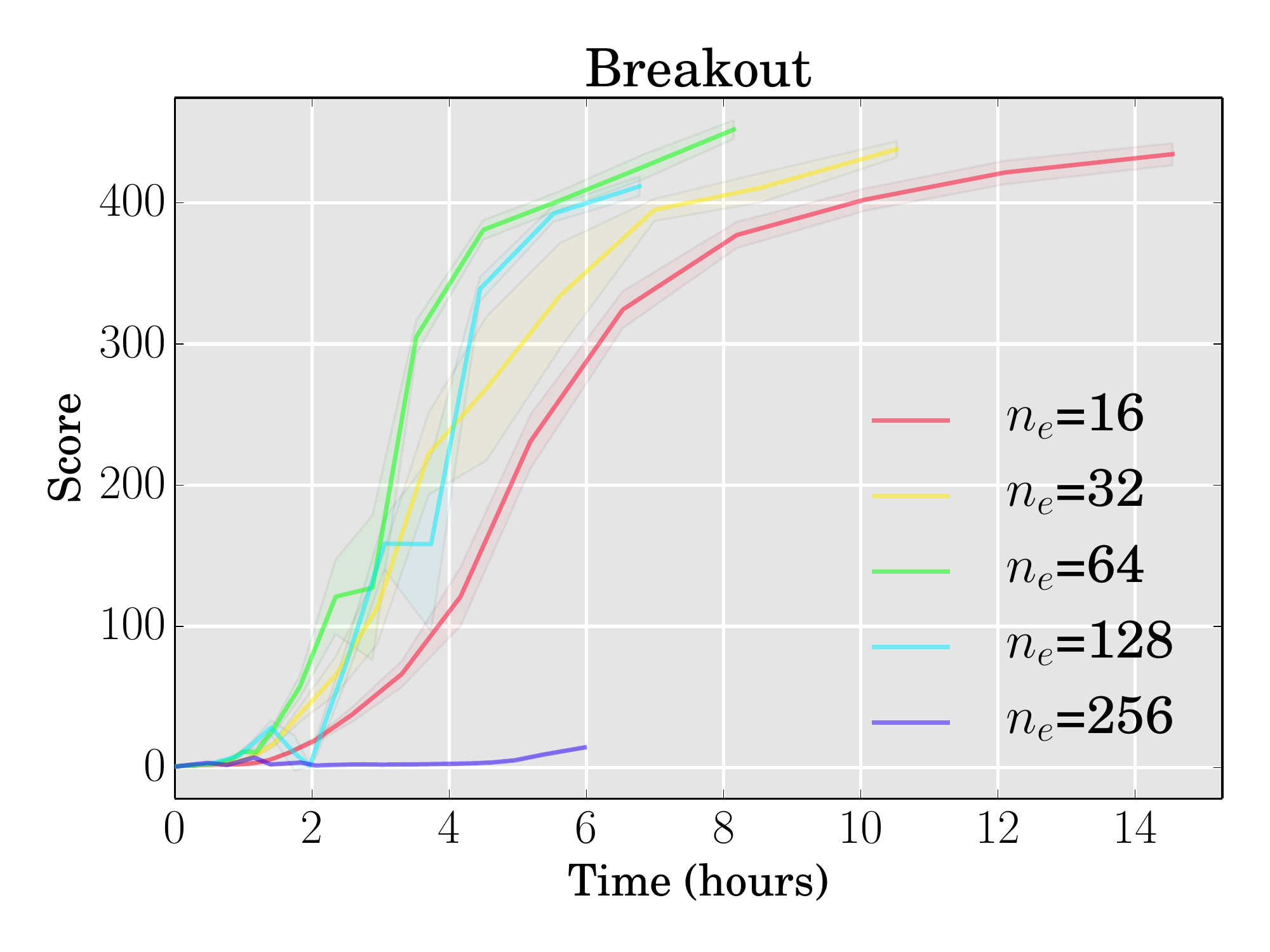}}        
    \mbox{\includegraphics[width=0.32\linewidth]{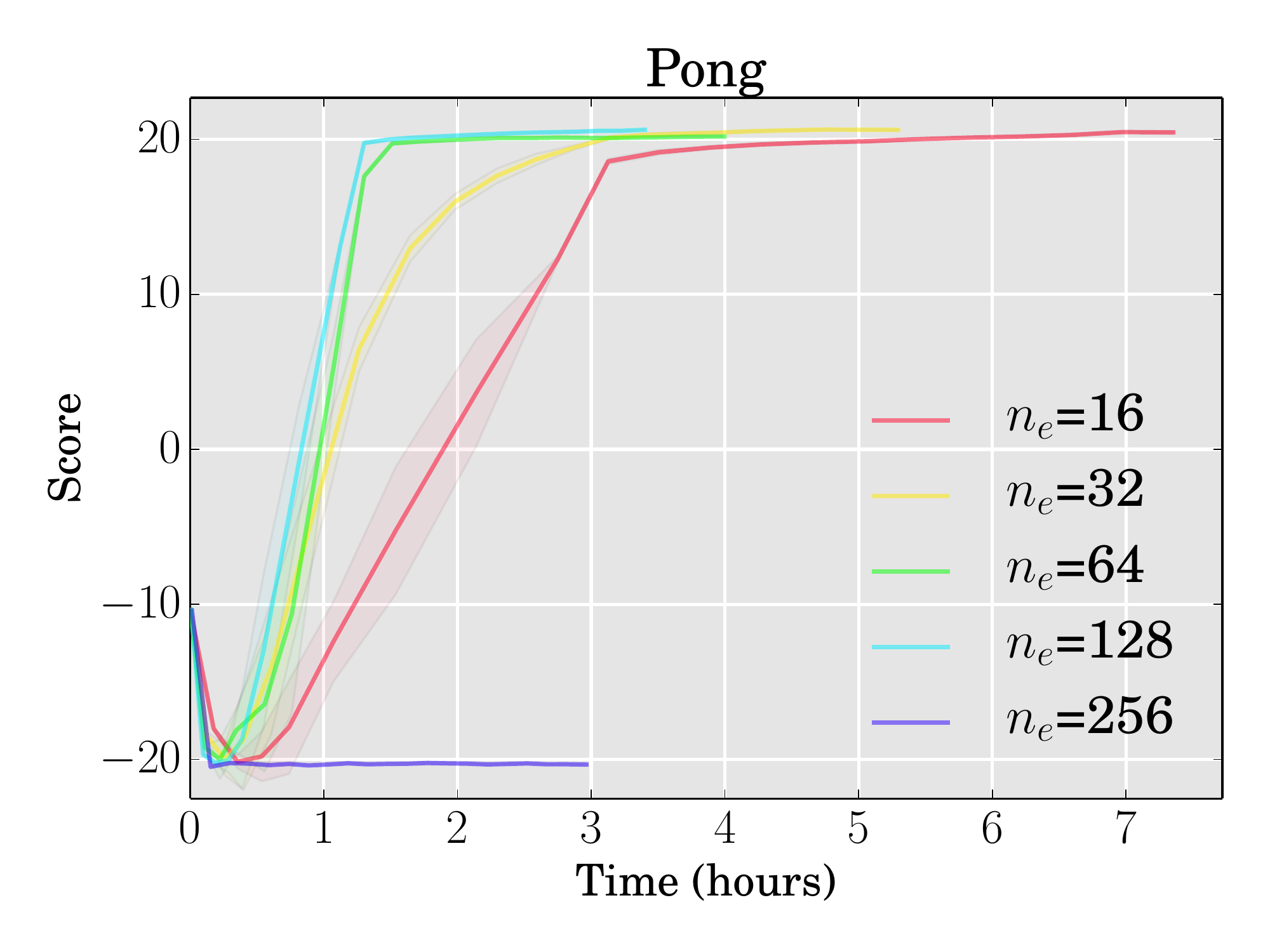}}   
    \hspace{2px}
    \mbox{\includegraphics[width=0.32\linewidth]{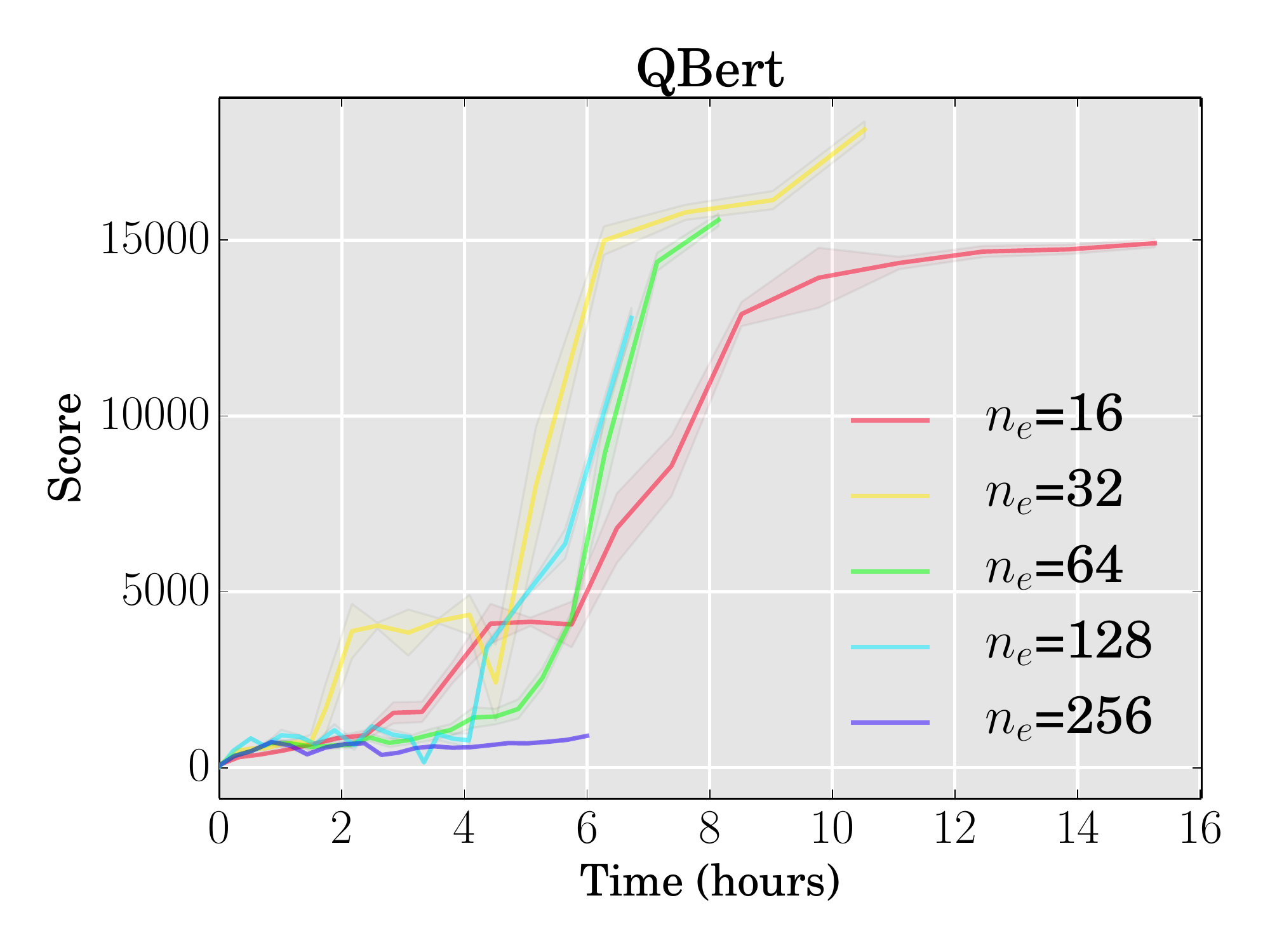}}
    \hspace{1px}
    \mbox{\includegraphics[width=0.32\linewidth]{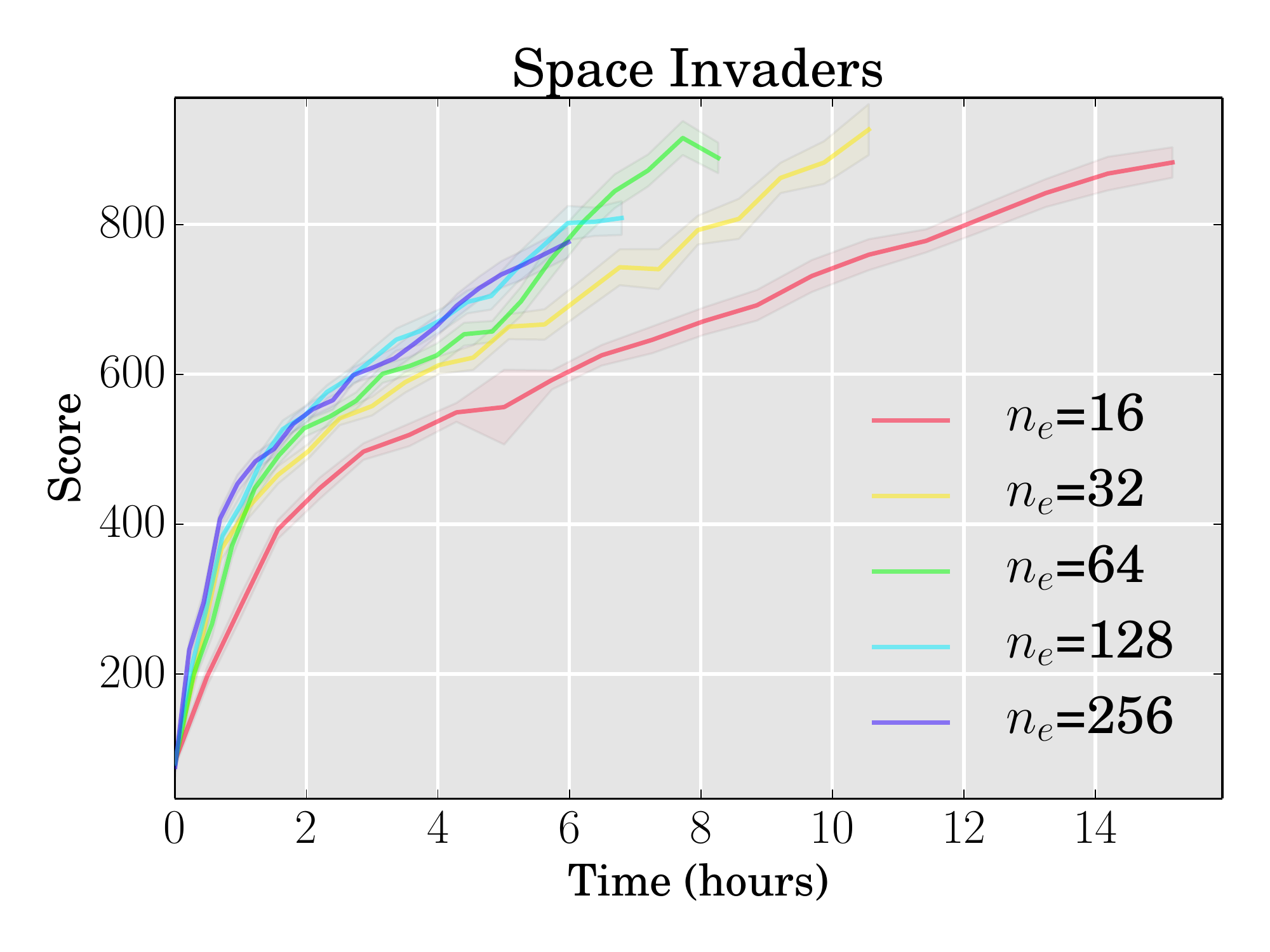}}
    \caption{Training time and score comparison for PAAC on six Atari 2600 games for different $n_e$.}
    \label{fig:training_time}
\end{figure}

\begin{table}[httb]
\centering
\scriptsize
\begin{tabular}{lcccccc}
\toprule
Game & Gorila & A3C FF & GA3C & PAAC $\text{arch}_{\text{nips}}$ & PAAC $\text{arch}_{\text{nature}}$\\
\cmidrule(lr){2-6}
Amidar & 1189.70 & 263.9 & 218 & 701.8 & 1348.3 \\
Centipede & 8432.30 & 3755.8 &  7386 & 5747.32 & 7368.1\\
Beam Rider & 3302.9 & 22707.9 & N/A & 4062.0 & 6844.0\\
Boxing & 94.9 & 59.8 & 92 & 99.6 & 99.8\\
Breakout & 402.2 & 681.9 & N/A & 470.1 & 565.3\\
Ms. Pacman & 3233.50 & 653.7 & 1978 & 2194.7 & 1976.0\\
Name This Game & 6182.16 & 10476.1& 5643 & 9743.7 & 14068.0\\
Pong & 18.3 & 5.6 & 18 & 20.6 & 20.9\\
Qbert & 10815.6 & 15148.8 & 14966.0 & 16561.7 & 17249.2\\
Seaquest & 13169.06 & 2355.4 & 1706 & 1754.0 & 1755.3 \\
Space Invaders & 1883.4 & 15730.5 & N/A & 1077.3 & 1427.8\\
Up n Down & 12561.58 & 74705.7 & 8623 & 88105.3 & 100523.3\\
\cmidrule(lr){1-6}
Training & 4d CPU cluster & 4d CPU & 1d GPU & 12h GPU & 15h GPU\\
\bottomrule
\end{tabular}
\caption{Scores are measured from the best per-
forming actor out of three, and averaged over 30
runs with upto 30 no-op actions start condition. Results for A3C FF use human start condition are therefore not directly comparable. Gorila scores taken from \citet{gorilla}, A3C FF scores taken from \citet{a3c} and GA3C scores take from \citet{nvidia}. Unavailable results are shown as N/A.}
\label{table:results}
\end{table}
\begin{figure}[ht]       
	\mbox{\includegraphics[width=0.32\linewidth]{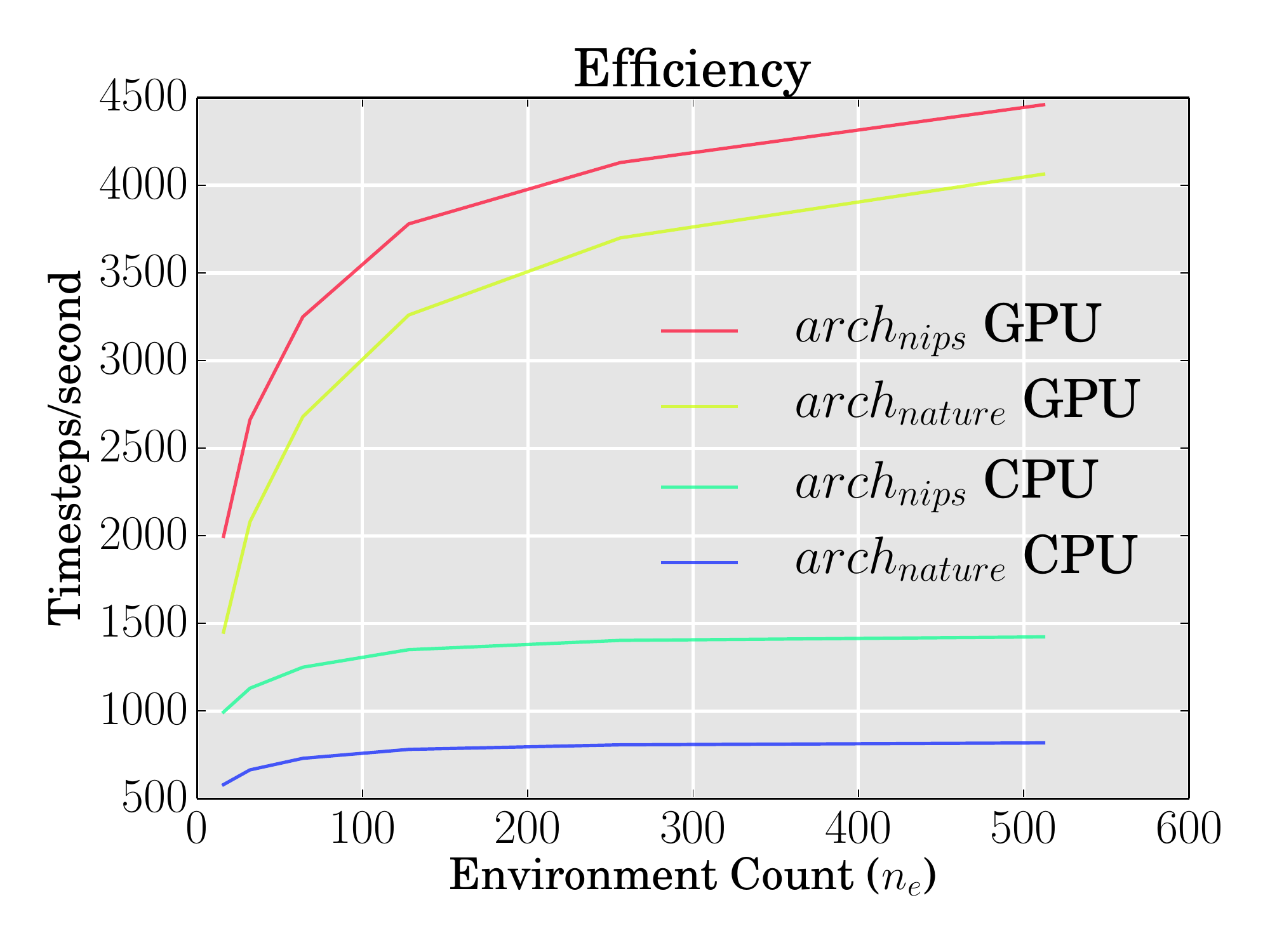}}   
    \hspace{1px}
    \mbox{\includegraphics[width=0.32\linewidth]{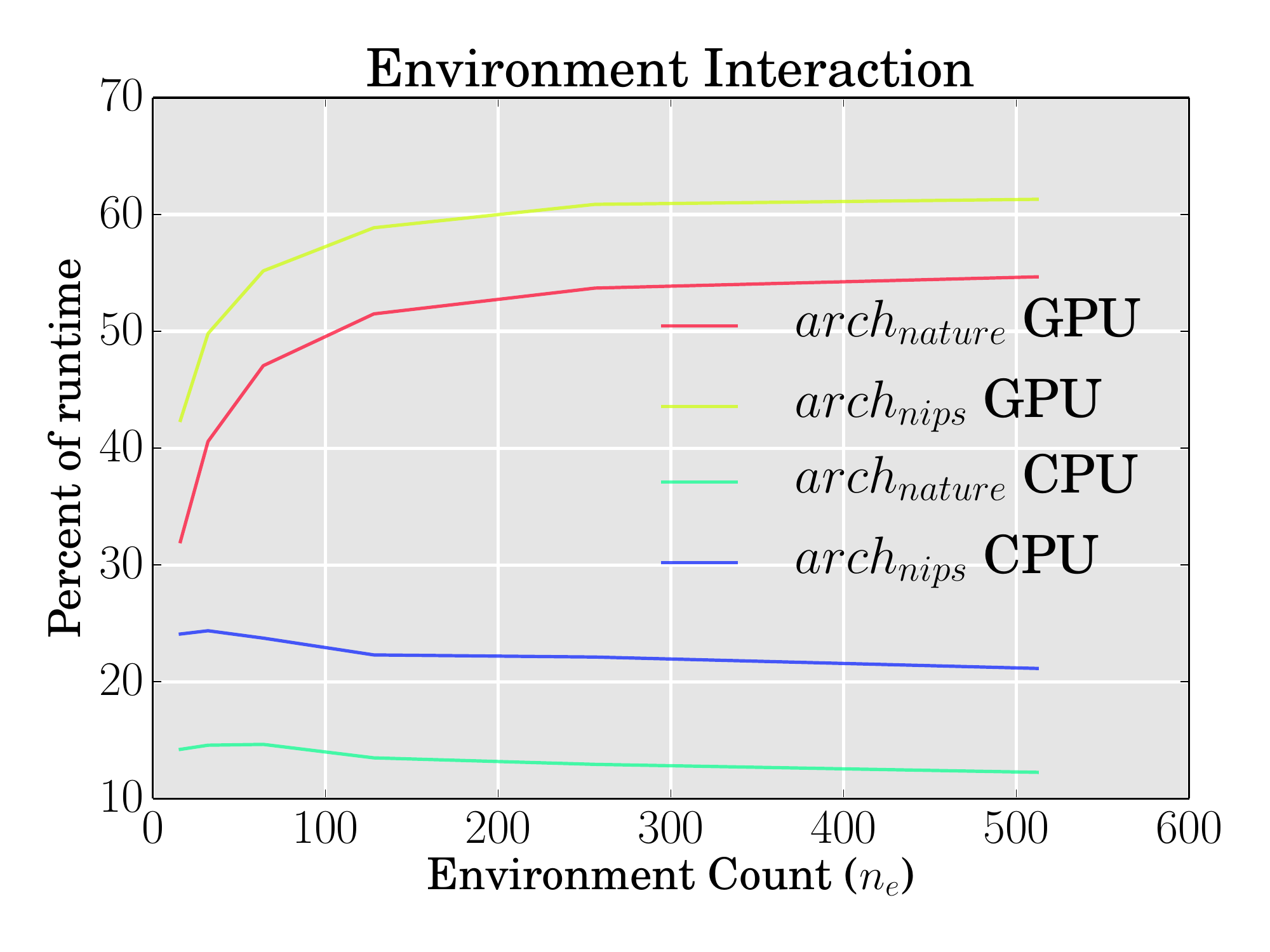}}
    \hspace{1px}
    \mbox{\includegraphics[width=0.32\linewidth]{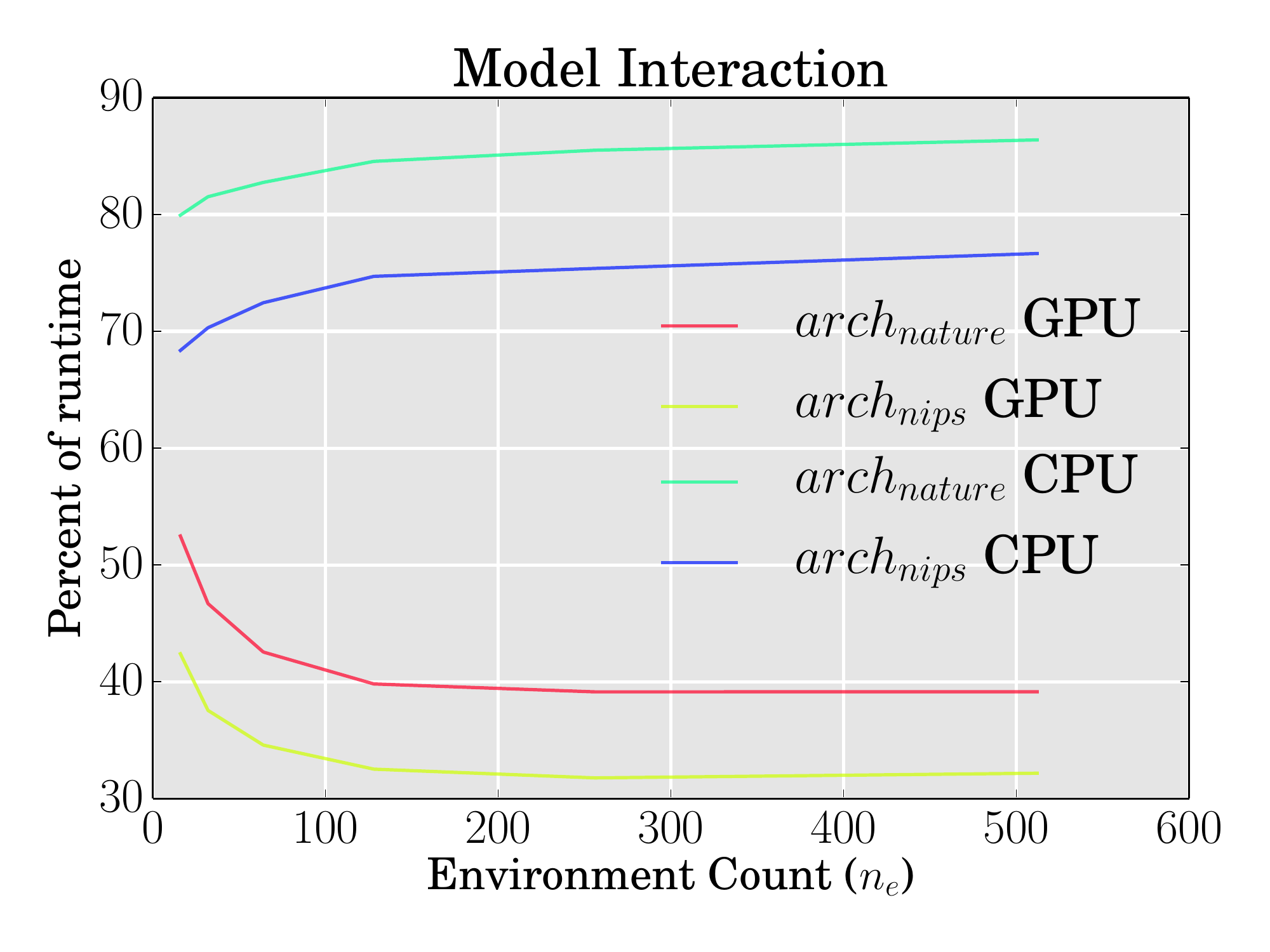}}        
    \caption{Time usage in the game of Pong for different $n_e$.}
    \label{fig:runtimes}
\end{figure}
To better understand the effect of the number of actors (and batch size) on the score, tests were run with $n_e \in \{16, 32, 64, 128, 256\}$. The learning rate was not tuned for each batch size, and was chosen to be $0.0007\cdot n_e$ for all runs across all games. Increasing $n_e$ decreases the frequency of parameter updates, given that parameter updates are performed every $n_e\cdot t_{\max}$ timesteps. As the theory suggests, the decreased frequency in parameter updates can be offset by increasing the learning rate. As can be seen in Figure~\ref{fig:score_comparison} most choices of $n_e$ result in similar scores at a given timestep, however Figure~\ref{fig:training_time} shows that higher values of $n_e$ reach those timesteps significantly faster. The choice of $n_e = 256$ results in divergence in three out of the four games, which shows that the learning rate can be increased proportional to the batch size, until a certain limit is reached. 
\vspace{-15px}
A limiting factor in the speed of training is the time spent in agent-environment interaction. When using $\text{arch}_{\text{nips}}$ for $n_e=32$ approximately 50\% of the time is spent interacting with the environment, while only 37\% is used for learning and action selection, as is shown in Figure~\ref{fig:runtimes}. This has strong implications for the models and environments that can be used. Using a model-environment combination that doubles the time needed for learning and action calculation would lead to a mere 37\% increase in training time. This can be seen in Figure~\ref{fig:runtimes} where using $\text{arch}_{\text{nature}}$ on the GPU leads to a drop in timesteps per second 22\% for $n_e=32$ when compared to $\text{arch}_{\text{nips}}$. When running on the CPU however this leads to a 41\% drop in timesteps per second.
\section{Conclusion}
In this work, we have introduced a parallel framework for deep reinforcement learning that can be efficiently parallelized on a GPU. The framework is flexible, and can be used for on-policy and off-policy, as well as value based and policy gradient based algorithms. The presented implementation of the framework is able to reduce training time for the Atari 2600 domain to a few hours, while maintaining state-of-the-art performance. Improvements in training time, will allow the application of these algorithms to more demanding environments, and the use of more powerful models.

\bibliography{iclr2016_conference}{}
\bibliographystyle{iclr2016_conference}
\end{document}